\pdfoutput=1

\documentclass[11pt]{article}

\usepackage{acl}

\usepackage{times}
\usepackage{latexsym}
\usepackage{amsmath}
\usepackage{subcaption}
\usepackage{hyperref}
\usepackage{pgfplots}
\usepackage{geometry}

\usepackage[T1]{fontenc}

\usepackage[utf8]{inputenc}

\usepackage{microtype}

\usepackage{inconsolata}

\usepackage{graphicx}

\title{SemCSE: Semantic Contrastive Sentence Embeddings Using LLM-Generated Summaries For Scientific Abstracts}

\author{Marc Brinner \\
  Computational Linguistics \\
  Bielefeld University, Germany \\
  {\tt marc.brinner@uni-bielefeld.de} \\\And
  Sina Zarrieß \\
  Computational Linguistics \\
  Bielefeld University, Germany \\
  {\tt sina.zarriess@uni-bielefeld.de} \\}
  \author{Marc Brinner \and Sina Zarrieß \\
  Computational Linguistics, Department of Linguistics\\
  Bielefeld University, Germany\\
  \texttt{\{marc.brinner,sina.zarriess\}@uni-bielefeld.de}}  

\begin{document}
\maketitle
\begin{abstract}

We introduce SemCSE, an unsupervised method for learning semantic embeddings of scientific texts. Building on recent advances in contrastive learning for text embeddings, our approach leverages LLM-generated summaries of scientific abstracts to train a model that positions semantically related summaries closer together in the embedding space. This resulting objective ensures that the model captures the true semantic content of a text, in contrast to traditional citation-based approaches that do not necessarily reflect semantic similarity. To validate this, we propose a novel benchmark designed to assess a model's ability to understand and encode the semantic content of scientific texts, demonstrating that our method enforces a stronger semantic separation within the embedding space. Additionally, we evaluate SemCSE on the comprehensive SciRepEval benchmark for scientific text embeddings, where it achieves state-of-the-art performance among models of its size, thus highlighting the benefits of a semantically focused training approach.

\end{abstract}

\section{Introduction}
\label{sec:intro}

The rapid growth in scientific publications \cite{bornmann_growth_2021} presents significant challenges for researchers in navigating the expanding body of knowledge. To address this, various embedding methods have been developed, both specifically for the scientific domain \cite{cohan2020specter, ostendorff_neighborhood_2022} and for text retrieval in general \cite{sturua2024jina, lee2025nvembed}. These methods transform texts into dense vector representations, enabling efficient assessment of semantic relatedness through vector comparison, thus supporting a range of downstream applications, including classification, clustering, and search \cite{subakti_performance_2022, singh_scirepeval_2023}, as well as modern applications like retrieval-augmented generation \cite{gao2024retrieval}.

The scientific domain in particular provides an exceptionally rich environment for both training and deploying embedding models, as, on the one hand, paper titles and abstracts are widely available and effectively encapsulate the core content of a publication, thus making them especially valuable for tasks like literature search and retrieval. On the other hand, this potential is further enhanced by the presence of citation links, which have long been recognized as a useful supervision signal indicating relatedness of scientific papers \cite{cohan2020specter, ostendorff_neighborhood_2022, mysore2022multi}.

While citations can serve as a useful proxy for semantic similarity, they introduce significant noise due to several factors, including 1) varying citation practices across disciplines \cite{hjorland_toward_1995}, 2) frequent citation of popular foundational works irrespective of their direct relevance, 3) interdisciplinary research including citations to fields with little thematic connection, and 4) citations made out of professional courtesy rather than genuine relatedness \cite{pasternack_scientific_1969}. Moreover, the absence of a citation does not necessarily indicate a lack of thematic overlap, as researchers may simply be unaware of each other's work.

To address these limitations, we propose SemCSE - a novel, fully unsupervised method for embedding scientific abstracts that emphasizes semantic content over external signals such as citation patterns. Our approach leverages a large language model to generate summarizing sentences that capture the core semantic information of scientific abstracts. These summaries are then used to train an embedding model to place summaries of the same abstract at nearby locations in the embedding space while pushing apart unrelated ones, thus effectively encouraging the model to learn robust and semantically meaningful representations of scientific texts.

A key advantage of our method is its unsupervised nature, which enables fast and scalable adaptation to new domains without the need for labeled data, thus contrasting supervised approaches that rely on large annotated datasets \cite{singh_scirepeval_2023} or large citation networks \cite{ostendorff_neighborhood_2022}.

A central contribution of our work is the paradigm shift from reliance on citation-based signals to a direct focus on semantic similarity. As existing benchmarks do not adequately capture this distinction, we introduce a novel benchmark specifically designed to assess a model's ability to encode the true semantic content of scientific texts. Our results show that SemCSE outperforms existing models trained using citation-based supervision, achieving a significantly stronger semantic separation in the embedding space. Furthermore, we validate the broader effectiveness of our approach by evaluating it on the comprehensive SciRepEval benchmark for scientific text representations \cite{singh_scirepeval_2023}, where SemCSE achieves state-of-the-art performance among models of comparable size.

\section{Related Work}
\label{sec:related_work}

\textbf{Structured Representations of Texts} are widely adopted for tasks like assessing semantic textual similarity \cite{li2024aoe}, question answering \cite{karpukhin2020dense}, document retrieval \cite{tang2021document} and clustering \cite{hadifar2019self}, and have been trained either using explicit supervision (e.g., \cite{reimers2019sentence}) or with unsupervised objectives (e.g., \citet{wu2020clearcontrastive}, \citet{gao2021simcse}, \citet{huang2021whiteningbert}).

\textbf{Scientific Document Embeddings} are a natural extension of this general development, and are commonly used to embed scientific papers for tasks like document retrieval \cite{kanakia_scalable_2019, NEURIPS2023_78f9c04b}, domain analysis and visualization \cite{lv_can_2024}, or as pretraining strategy for creating domain-specific transformer models \cite{brinner2025enhancingdomainspecificencodermodels}. While simple methods leverage basic word-frequency information \cite{achakulvisut_science_2016, meijer_document_2021}, recent approaches train neural network embedding models, for example by using an unsupervised contrastive objective that enforces similar embeddings for different parts of the same document \cite{tan_self-supervised_2023}, or by using explicit supervision in the form of classification and regression tasks \cite{singh_scirepeval_2023}.

In contrast to these examples, most embedding models for scientific texts rely on citation relationships as a proxy for semantic relatedness between papers, thus enforcing similar embeddings for papers that share a citation link. \citet{bhagavatula_content-based_2018} use this to train text representations based on weighted word vectors, while \citet{cohan_specter_2020} use the same concept for training a transformer embedding model, with \citet{ostendorff_neighborhood_2022} improving the selection of negative samples using a citation network embedding space.

More recent developments have focused on creating \textbf{Task-Specific Embeddings}, thus creating multiple embeddings for a given document encoding different aspects, or creating embeddings specifically suitable for certain tasks \cite{mysore2022multi, singh_scirepeval_2023, lee2025nvembed}.

A different line of research instead focuses on leveraging \textbf{Synthetic Data For Embedding Models}, which was proven to be effective both for use during training \cite{lee2024geckoversatiletextembeddings, wang-etal-2024-improving-text, chen-etal-2025-little} or inference \cite{frank2024gasegenerativelyaugmentedsentence, thirukovalluru-dhingra-2025-geneol}. Notably, the use of LLM-generated synthetic data for training embedding models remains unexplored in the scientific domain. Further, existing research focuses on obtaining high-quality training data, usually by using large proprietary LLMs, which contrasts our method that proves to be effective by leveraging small LLMs as tools for simple semantics-preserving data augmentation.

\section{Method}
\label{sec:method}

We propose SemCSE, a simple contrastive learning scheme designed to train a text embedding model with a strong emphasis on accurate semantic representation. While our experiments focus on the scientific domain, we believe our approach is broadly applicable. Therefore, we present the method in a general form here and provide domain-specific details and adaptations in Section \ref{sec:experiments}.

Our embedding approach is based on a dataset of texts representing the target domain, denoted as $\mathcal{A} = \{A_1, A_2, ..., A_n\}$. Using this dataset, the first step in our training pipeline involves using an LLM to generate multiple summarizing sentences for each text in the training set, resulting in a dataset $\mathcal{S} = \{(s_{1, 1}, s_{1, 2}, ...), (s_{2, 1}, s_{2, 2}, ...), ..., (s_{n, 1}, s_{n, 2}, ...)\}$. This dataset is subsequently used to train the embedding model via a triplet loss, encouraging summaries of the same text to be mapped to nearby locations in the embedding space.

Formally, for each index $i_j$ within a batch $\mathcal{B} = \{i_1, ..., i_{|\mathcal{B}|}\}$ of sampled indices, we randomly select two summarizing sentences for the corresponding text $T_{i_j}$ and denote them as $s_{j,1}$ and $s_{j,2}$. We then embed them individually using our model $M$:
\begin{align*}
    e_{j, 1} = M(s_{j, 1}) \\
    e_{j, 2} = M(s_{j, 2}) 
\end{align*}
For each pair of matching summaries, we define $e_{j, 1}$ as the \textit{anchor} $e_a$ and $e_{j, 2}$ as the \textit{positive} $e_+$, and sample a third, random, summary as \textit{negative} $e_-$. On these triples, we compute the following triplet loss \cite{triplet_loss}:
\begin{align*}
    \mathcal{L}(e_{\textrm{a}}, e_+, e_-) = \textrm{relu}(d(e_a, e_+) - d(e_a, e_-) + 1)
\end{align*}
Here, $m$ is a margin hyperparameter, and $d$ is a distance function (e.g., Euclidean distance, which is used in our experiments). This loss encourages the model to embed the anchor and positive at least one unit closer together than the distance between the anchor and negative. Thus, the model learns to create embeddings that capture the semantic content of each sentence to ensure that semantically similar summaries are positioned close together in the embedding space. The final loss for the entire batch is then formulated as:
\begin{align*}
    \mathcal{L} = \frac{1}{|\mathcal{B}|}\sum_{i \in \mathcal{B}} \frac{1}{|\mathcal{B}|-1} \sum_{j \in B, j \neq i} \mathcal{L}(e_{i, 1}, e_{i, 2}, e_{j, 2})
\end{align*}
This formulation creates $|\mathcal{B}|-1$ triples for each positive pair by selecting each positive from other pairs as negative. This improves training significantly since, as the model improves, many triples will yield a zero loss and incorporating a larger number of triples increases the likelihood of obtaining informative gradient signals.

In addition to the base objective, we apply a weak L2 regularization to the embeddings to encourage a more compact embedding space.

\subsection{A Comparative Analysis}
\label{sec:comp_analysis}

Contrastive loss formulations have proven highly effective for training embedding models, with key insights from \citet{wang2022understand} being that their success largely stems from promoting uniformity - i.e., encouraging embeddings to be evenly distributed in the embedding space to allow for better disambiguation - as well as from promoting alignment, meaning that semantically similar inputs are placed close together.

While alignment is typically enforced by using positive pairs from supervised datasets, \citet{gao2021simcse} created an unsupervised contrastive loss by using the same sentence as both the anchor and the positive, thus mainly focusing on improving uniformity through pushing unrelated samples apart. Notably, they show that introducing variance between the anchor and positive embeddings - enabled through dropout in the forward pass - is key for maintaining alignment, which is otherwise not promoted due to the lack of related positive pairs.

Our own experiments support this analysis, since we observed that model performance using unsupervised SimCSE peaked after about 1,000 steps, suggesting that the training process quickly saturates if uniformity is sufficiently improved, and that alignment is not further promoted through meaningful information from related samples.

In contrast, our proposed method introduces a more semantically informative training signal by using related but clearly distinct summarizing sentences created by leveraging the stochasticity of autoregressive generation. These summaries serve as anchor-positive pairs, thus presenting a more challenging learning task that requires the model to identify the shared meaning across diverse surface forms, ultimately preserving alignment within this broader space of semantically related sentences.

It is important to note that the concepts of alignment and uniformity, as defined by \citet{wang2022understand}, are formally grounded in a hyperspherical embedding space induced by the use of cosine similarity and thus do not directly transfer to the unconstrained Euclidean space employed in our study. Nevertheless, the underlying principles of encouraging the separation of unrelated embeddings for easier disambiguation and bringing related inputs closer together remain conceptually applicable. For this reason, we perform an empirical analysis of the effect of our training scheme on anisotropy of the embedding space in Appendix \ref{sec:anisotropy}.

\section{Model Training}
\label{sec:experiments}

We investigate the applicability of our training scheme to scientific texts, thus creating an embedding model suitable for tasks like literature search.

\begin{figure*}
    \centering
    \begin{subfigure}[b]{0.32\linewidth}
        \centering
        \fbox{\includegraphics[width=\linewidth]{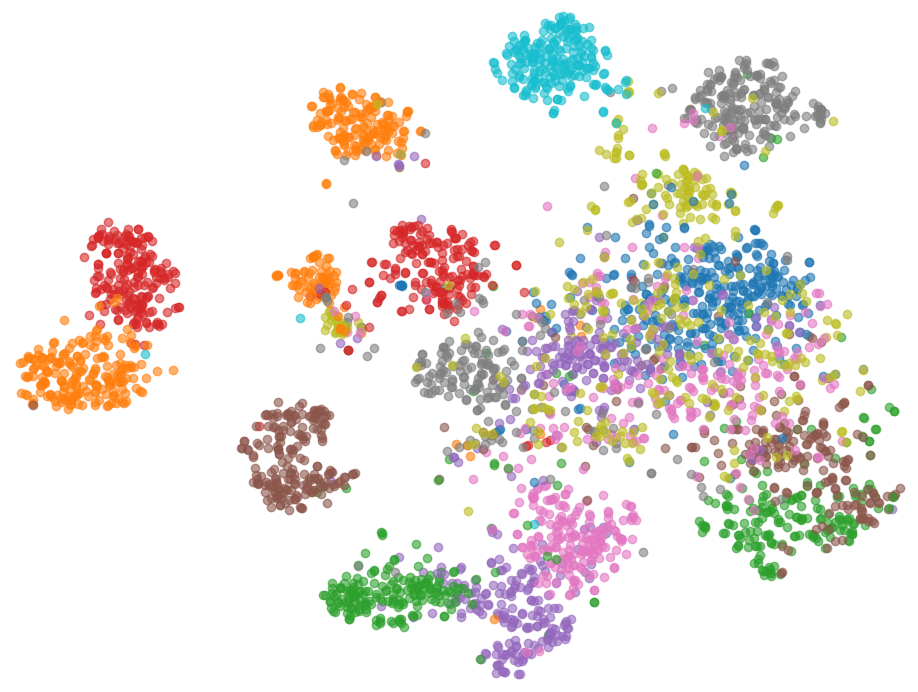}}
        \caption*{\large SemCSE}
    \end{subfigure}
    \hfill
    \begin{subfigure}[b]{0.32\linewidth}
        \centering
        \fbox{\includegraphics[width=\linewidth]{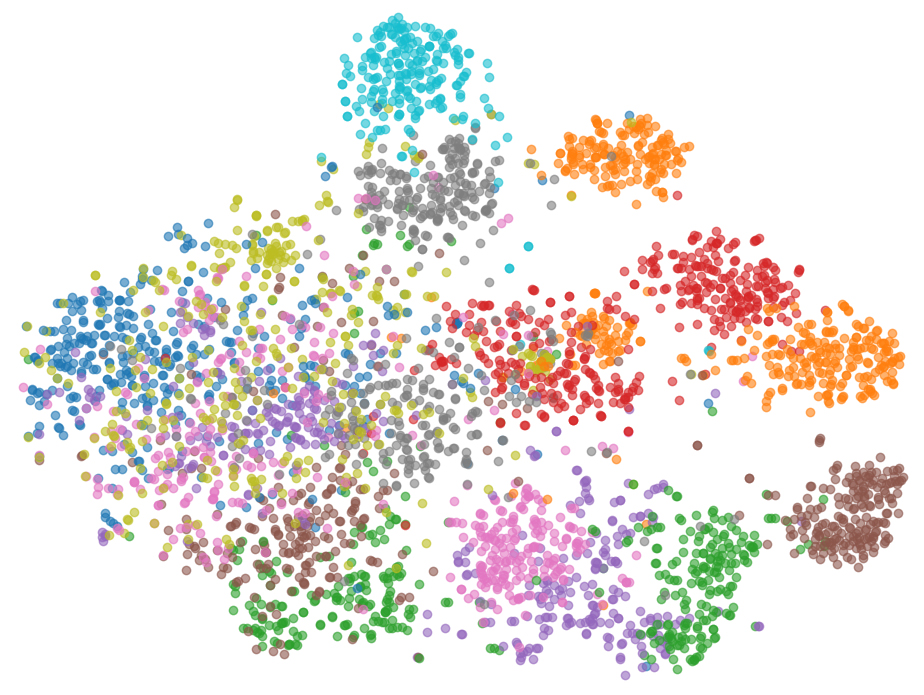}}
        \caption*{\large SPECTER2 base}
    \end{subfigure}
    \hfill
    \begin{subfigure}[b]{0.318\linewidth}
        \centering
        \fbox{\includegraphics[width=\linewidth]{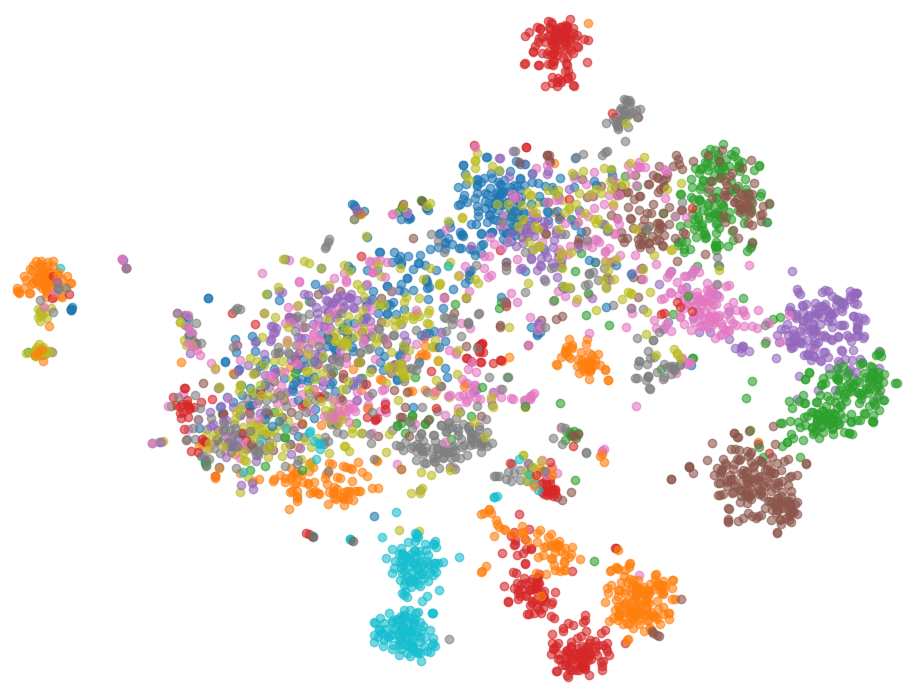}}
        \caption*{\large SemCSE - Same Input}
    \end{subfigure}
    \vspace{-5px}
    \caption{A t-SNE visualization of embeddings generated for scientific papers from the SciDocs MAG dataset \cite{cohan_specter_2020}, with points colored according to their assigned topic labels.}
    \label{fig:embeddings}
    \vspace{-8px}
\end{figure*}

\subsection{Dataset}

As training data, we utilize a collection of corpora from the SciRepEval benchmark \cite{singh_scirepeval_2023}, a large-scale evaluation suite for scientific text embedding models. From these corpora, we sample 350K paper titles and abstracts spanning a variety of domains (for details, see Appendix \ref{sec:appendix}).

Our method requires multiple summarizing sentences per sample, which we generate by concatenating title and abstract and processing them with Llama-3-8B \cite{grattafiori2024llama3herdmodels}. Specifically, we append one of five predefined prompts (e.g., "A comprehensive summary for our work would be that") to the abstract and generate three continuations, thus effectively summarizing the abstract. To create a more challenging matching task, these prompts are designed to extract different types of information, including the general topic of the research, comprehensive summaries or key findings.

We also performed preliminary experiments with summaries created by chat LLMs, but found that these performed slightly worse. We hypothesize that the continuation-based approach adheres more closely to the input data distribution, generating sentences that more naturally resemble those found in scientific abstracts. A full list of prompts and other training details are provided in Appendix \ref{sec:data_generation}, while a discussion about the effect of summary quality is presented in Appendix \ref{sec:summary_quality}.

\subsection{Model Training}
\label{sec:model_training}
As base model within our experiments, we use a pretrained SciDeBERTa model \cite{10355927}.

While we only use generated summaries as anchors, we increase variance within the positives by also sampling paper titles or sentences from the abstracts in 15\% and 35\% of cases, respectively. This forces the model to learn meaningful relationships between different representations of the same document, leading to a deeper semantic understanding. This alteration is not applied to the anchors to ensure that one representation of the text captures the underlying semantics comprehensively.

\subsection{Baselines}

For our evaluation we compare SemCSE to a diverse set of embedding models, including those specifically designed for scientific texts in the form of SciBERT \cite{beltagy_scibert_2019}, SciDeBERTa \cite{10355927}, SPECTER \cite{cohan2020specter}, SciNCL \cite{ostendorff_neighborhood_2022}, and SPECTER2 \cite{singh_scirepeval_2023}, as well as several state-of-the-art general-purpose embedding models commonly used for document retrieval. These include all-MiniLM-L6-v2\footnote{\url{https://huggingface.co/sentence-transformers/all-MiniLM-L6-v2}}, jina-embeddings-v2-base-en \cite{günther2023jina}, jina-embeddings-v3 \cite{sturua2024jina}, NvEmbed-V2 \cite{lee2025nvembed}, and RoBERTa-SimCSE \cite{gao2021simcse}.

Since we focus on evaluating a model's ability to create generally applicable and task-independent semantic embeddings, we do not use domain-specific prompts that allow for task-specific embeddings (e.g., for NvEmbed), or task-specific adapters (e.g., for SPECTER2), and instead rely on the component for general semantic embedding.

\section{Evaluating Semantic Embedding Capabilities}
\label{sec:semantic_eval}

\subsection{Generalization to Longer Texts}
\label{sec:generalization}

\begin{table*}[ht!]
  \centering
  \resizebox{\textwidth}{!}{
  \begin{tabular}{l|c|c|c|c|c|c}
    \hline
    \textbf{Model} & Parameters & Title-Abstr. $\downarrow$ & Abstr.-Segments $\downarrow$ & Query $\downarrow$ & Clustering $\uparrow$ & Perf. $\uparrow$ \\
    \hline
    SciBERT & 109M & 807.74 & 214.37 & 213.45 & 0.569 & 0.000 \\
    SciDeBERTa & 183M & 1479.09 & 861.55 & 2465.26 & 0.460 & 0.000 \\
    SPECTER & 109M & 10.25 & 12.23 & 2.18 & 0.692 & 0.119 \\
    SciNCL & 109M & 5.68 & 7.35 & 2.29 & 0.702 & 0.357 \\
    SPECTER2 \small base & 109M & 4.52 & 5.10 & 1.17 & 0.666 & 0.553 \\
    SPECTER2 \small proximity & 110M & 5.34 & 5.80 & 1.46 & 0.666 & 0.395 \\ \hline
    all-MiniLM-L6-v2 & 22M & \underline{3.09} & 8.19 & 1.11 & \underline{0.730} & 0.771 \\
    Jina-v2 & 137M & 3.29 & 8.77 & 1.29 & 0.703 & 0.600 \\
    Jina-v3 & 572M & 3.45 & 6.96 & \textbf{1.01} & 0.719 & 0.783 \\
    RoBERTa SimCSE & 355M & 23.71 & 44.24 & 8.92 & 0.696 & 0.116 \\
    NvEmbed-V2 & 7.9B & 3.38 & \underline{3.84} & \underline{1.02} & 0.721 & \underline{0.866} \\ \hline
    SemCSE (Ours) & 183M & \textbf{2.47} & \textbf{2.68} & 1.23 & \textbf{0.739} & \textbf{0.925} \\ \hline
  \end{tabular}}
  \caption{Results for evaluating SemCSE and baseline models on the semantic embedding benchmark. The best scores are bold, while second-best are underlined.}
  \label{tab:results_semantic}
\end{table*}

Since our model is trained exclusively on individual sentences - i.e., summaries, paper titles, or randomly sampled sentences from abstracts - it is essential to evaluate its ability to generalize to longer texts and capture their overall semantics.

To this end, we consider the validation dataset comprising 900 title-abstract pairs and corresponding paper summaries. We assess performance using a ranking-based retrieval metric: For a given summary, the model must identify the matching title-abstract pair from a pool of 900 candidates based on embedding proximity. We report the average rank at which the correct match is retrieved, with 1 being optimal and 900 the worst.

Our model achieves an average rank of 1.542, indicating a capability to produce semantically meaningful embeddings for both short and long texts that supports precise semantic retrieval. When using only the title or only the abstract instead of their concatenation, performance drops to 3.0 and 1.801, respectively. These results confirm that the model does not rely solely on the opening sentence and is capable of effectively embedding longer and more complex texts than those seen during training. Moreover, the results highlight the model's ability to embed diverse forms of scientific text - summaries, titles, and abstracts - into a unified semantic space.

\subsection{Semantic Embedding Benchmark}

Building on the analyses from the previous section, we introduce a benchmark specifically designed to evaluate a model’s ability to capture the precise semantic content of scientific texts. While existing benchmarks such as SciRepEval (Section \ref{sec:scirepeval}) include tasks like citation prediction, same-author identification, and citation count estimation, these tasks often do not evaluate a model's semantic embedding capabilities. For example, citation-based links do not necessarily imply semantic similarity (see Section \ref{sec:intro}), authors may shift research topics over time, and citation counts can vary for reasons unrelated to a paper's content

To address these limitations, we propose four tasks that more effectively evaluate a model’s semantic embedding capabilities for scientific texts.

The \textbf{Title–Abstract Matching} task measures a model’s ability to match a paper’s title with its corresponding abstract. Titles and abstracts both serve as compressed representations of the same underlying work, albeit at different levels of detail, so that a model that captures semantic meaning should predict similar embeddings for both of these representations.

The \textbf{Abstract Segmentation Consistency} task tests the model’s ability to match two halves of the same abstract. Given that scientific abstracts are highly condensed summaries of research papers, both halves should contain considerable information about the core theme of the paper, and a model that effectively captures this information should again predict similar embeddings.

The \textbf{Query Matching} task evaluates a model’s ability to associate a scientific paper with a relevant search query. In this case, we pair each title-abstract pair with a search query generated using Mistral Small 3.1 \cite{mistral}, a state-of-the-art 27B-parameter LLM. Despite surface-level differences, a model with strong semantic understanding should assign similar embeddings to both the query and the corresponding paper.

Similar evaluation strategies have already been proven successful in the context of passage retrieval \cite{vasilyev-etal-2025-preserving}. To evaluate these tasks, we create a dataset consisting of 6,000 samples drawn from 12 datasets of the SciRepEval benchmark (details in Appendix \ref{sec:appendix}), and evaluate them using the ranking-based metric introduced in the previous section, thus reporting the average rank at which the correct match is retrieved from the candidate pool of 6,000 samples.

As a fourth task, we propose the \textbf{Semantic Clustering} task, which uses the SciDocs MAG dataset \cite{cohan2020specter}, containing 21,252 paper titles and abstracts annotated with thematic categories from the Microsoft Academic Graph. While the original benchmark proposed by \citet{cohan2020specter} focused on training a linear SVM to predict these categories from the embeddings, we argue that semantic similarity might not manifest in linear separability. Instead, we assess whether a model’s embedding space naturally clusters semantically related papers. To do this, we embed each paper (title + abstract) from both the train and test sets, and - for each test sample - retrieve its five nearest neighbors from the training set. We report the proportion of these neighbors that share the same thematic category, providing a measure of how well the model organizes scientific knowledge in a semantically meaningful structure.

Finally, we compute an overall performance score by normalizing task-specific results to a scale from 0 to 1 and averaging across all tasks. For each task, the best-performing model is assigned a score of 1, the median-performing model a score of 0.5, and intermediate values are linearly interpolated, with 0 being a threshold at the lower end.


\subsection{Results}

\begin{table*}
  \centering
  \begin{tabular}{l|c|c|c|c|c|c}
    \hline
    \textbf{Model} & Parameters & Classification & Regression & Proximity & Search & Average \\
    \hline
    SciBERT & 109M & 63.86 & 27.34 & 66.25 & 68.19 & 57.42 \\
    SciDeBERTa & 183M & 60.99 & 27.00 & 62.74 & 67.83 & 55.18 \\
    SPECTER & 109M & 67.73 & 25.37 & 80.05 & 74.89 & 64.28 \\
    SciNCL & 109M &  \underline{68.04} & 25.22 &  \underline{81.18} & 77.32 & 65.08 \\
    SPECTER2 base & 109 & 66.95 & \underline{27.75} & 81.10 & 78.42 & 65.46 \\
    SPECTER2 proximity & 110 & 66.37 & 26.85 & \textbf{81.41} & 77.75 & 65.15 \\ \hline
    all-MiniLM-L6-v2 & 22M & 64.04 & 20.06 & 80.74 & 79.63 & 63.05 \\
    jina-v2 & 137M & 63.99 & 23.76 & 80.11 & 80.40 & 63.69 \\
    jina-v3 & 572M & 65.66 & 24.84 & 79.98 &  \underline{80.60} & 64.34 \\
    RoBERTa SimCSE & 355M & 67.16 & 22.95 & 75.51 & 76.97 & 62.10 \\
    NvEmbed-V2 & 7.9B & 65.62 & \textbf{29.94} & 81.16 & \textbf{82.84} & \textbf{66.19} \\ \hline
    SemCSE (Ours) & 183M & \textbf{69.52} & 27.58 & 80.21 & 78.56 & \underline{65.76} \\ \hline
  \end{tabular}
  \caption{Results for evaluating SemCSE and baseline models on the SciRepEval benchmark. The best scores are bold, while second-best are underlined.}
  \label{tab:results_scirepeval}
  \vspace{-5px}
\end{table*}

The results of our semantic embedding evaluation benchmark are presented in Table \ref{tab:results_semantic}.

Examining the title-abstract and abstract-segments matching tasks, we observe that models not explicitly trained as embedding models (i.e., SciBERT and SciDeBERTa) struggle to accurately encode the semantics of a given paper title or abstract. While training on basic citation triples substantially improves performance (SPECTER, 10.25 and 12.23), scores remain significantly higher than those achieved by models employing more advanced training strategies, for example by leveraging improved negative sampling (SciNCL, 5.68 and 7.35) or integrating supervised proximity-based datasets (SPECTER2-proximity, 5.34 and 5.80).

Interestingly, general-domain retrieval models such as Jina-v2, Jina-v3, and even the small MiniLM model, excel at the title-abstract matching task - likely caused by the close resemblance to document retrieval in general search settings - but struggle on the less familiar abstract segments matching task.

Our SemCSE model achieves state-of-the-art performance in both title-abstract matching and abstract-segments matching despite being trained solely on individual sentences, even surpassing the powerful NvEmbed-V2 model that leverages more than 43 times as many parameters. We interpret the especially unrivaled performance on the abstract-segments task as evidence of a deepened understanding of scientific texts, since each half of an abstract lacks crucial information, so that a strong performance on this task requires recovering a precise semantic representation from reduced context.

The query matching task shows improved performance compared to the other matching tasks across most models, suggesting that the LLM-generated queries are well-aligned with the semantic content of their corresponding papers, thus avoiding the challenges posed by ambiguous titles and varying abstracts. As a result, retrieval-optimized models such as Jina-v3 and NvEmbed achieve near-perfect performance, with average ranks of 1.01 and 1.02, respectively. Our SemCSE model also performs well, achieving an average rank of 1.23.

In the semantic clustering task, our SemCSE model achieves a state-of-the-art score of 0.739, outperforming all other evaluated models. The closest competitor is MiniLM (0.730), followed by the significantly larger NvEmbed model (0.721). All remaining models - including those specifically trained on scientific literature - score considerably lower. We hypothesize that this is caused by their reliance on citation triples as a supervisory signal, which might link semantically unrelated papers from different domains, ultimately leading to a less semantically-separated embedding space. This is supported by the t-SNE visualization in Figure \ref{fig:embeddings}, where embeddings produced by SemCSE exhibit clearly separated thematic clusters, with SPECTER2-base (the best citation-based model on the SciRepEval benchmark) showing substantially weaker topic separation.

Evaluating the overall performance score, we see a clear lead by SemCSE (0.925), with only the 7.9B parameter NvEmbed model coming remotely close (0.866), thus again highlighting the strong semantic embedding capabilities of our approach.

\section{SciRepEval Evaluation}
\label{sec:scirepeval}

We further evaluate our model using the SciRepEval benchmark \cite{singh_scirepeval_2023}, which comprises 24 tasks designed to assess the performance of text embeddings across ad-hoc search, proximity, regression, and classification tasks. The task-type-aggregated results for all models are presented in Table \ref{tab:results_scirepeval}, with all individual results being included in Appendix \ref{app:scirep_individual}.

\subsection{Average Performance}

As seen in the context of the semantic benchmark, domain-specific models not explicitly trained for embedding tasks exhibit below-average performance. In contrast, training on citation data leads to strong overall scores, with models utilizing simple contrastive loss formulations (e.g., SPECTER, Average: 64.28) again underperforming compared to those employing more advanced training strategies, such as SciNCL (Average: 65.08).


While general-domain embedding models show strong results on some semantic matching tasks, their performance on SciRepEval is markedly lower - unsurprising given that many tasks in this benchmark are closely aligned with citation-based training signals used by domain-specific models. A notable exception is NvEmbed-V2, which achieves a state-of-the-art average score of 66.19, albeit at the cost of significantly higher computational cost.


The SemCSE model achieves the second-best overall score (65.76), outperforming other domain-specific models despite not being trained with the same citation-based supervision signals. This result strongly supports the validity of our method and further demonstrates that a greater emphasis on semantic representations can be highly beneficial.

\subsection{Task-Specific Performance}

Beyond aggregate scores, task-type-specific performance sheds light on the relative strengths of different pretraining approaches.

Our SemCSE model performs exceptionally well on classification tasks, achieving the highest score across all models. This is expected, as such tasks benefit most from semantically rich embeddings that facilitate clear class separation.

Regression tasks exhibit more varied outcomes across models, suggesting that no single training strategy is uniquely optimized for them.

Proximity-based tasks - many of which rely on citation information as ground truth - naturally favor models trained with citation-based supervision. Remarkably, despite not using such signals, SemCSE achieves strong performance in this category, suggesting that - although some citations may link semantically unrelated papers - citation links still broadly correlate with semantic similarity.

Finally, as expected, retrieval-optimized models outperform others on ad-hoc search tasks, even within the scientific domain.

\section{Ablation}
\label{sec:ablation}

\begin{table*}[ht!]
  \centering
  \resizebox{\textwidth}{!}{
  \begin{tabular}{l|c|c|c|c|c|c|c|c|c|c}
    \hline
    \textbf{Model} & Dataset & Clf. & Regr. & Prox. & Search & SRE Avg. & Title-Abstr. & Abstr.-Segments & Query & Clustering \\
    \hline
    Full & 350K & 69.52 & 27.58 & 80.21 & 78.56 & 65.76 & 2.47 & 2.68 & 1.23 & 0.739 \\
    Full & 175K & 69.26 & 27.19 & 80.36 & 78.54 & 65.67 & 3.40 & 3.21 & 1.29 & 0.732 \\
    Full & 87.5K & 69.01 & 27.39 & 80.35 & 78.58 & 65.65 & 3.35 & 2.63 & 1.24 & 0.731 \\
    Full & 35K & 69.32 & 27.18 & 80.18 & 77.81 & 65.51 & 3.65 & 3.24 & 1.24 & 0.734 \\
    Full & 17.5K & 68.31 & 27.00 & 80.04 & 78.37 & 65.23 & 3.76 & 3.47 & 1.45 & 0.723 \\
    Full & 8.75K & 68.95 & 27.05 & 79.96 & 76.53 & 65.14 & 4.34 & 3.73 & 1.99 & 0.725 \\
    Full & 3.5K & 68.36 & 27.07 & 79.94 & 76.73 & 65.01 & 5.93 & 4.04 & 2.18 & 0.728 \\ \hline
    Just Summaries & 350K & 68.78 & 26.89 & 80.21 & 77.51 & 65.28 & 6.04 & 3.52 & 2.11 & 0.732 \\
    Same Input & 350K & 66.59 & 27.23 & 74.29 & 69.94 & 61.48 & 180.01 & 59.86 & 251.46 & 0.645 \\ 
    Cosine Similarity & 350K & 69.76 & 25.88 & 80.63 & 79.33 & 65.72 & 2.84 & 3.17 & 1.13 & 0.735 \\ \hline
  \end{tabular}
  }
  \vspace{-3px}
  \caption{Results for SemCSE trained with different dataset sizes and variations of the loss function.}
  \label{tab:results_ablation}
  \vspace{-5px}
\end{table*}

We conduct several ablations to evaluate the robustness and effectiveness of our training strategy. The corresponding results are presented in Figure \ref{tab:results_ablation}.

We begin by assessing the impact of reduced training data. On the SciRepEval benchmark, performance remains relatively stable, maintaining scores above 65.0 even when using only 1\% of the original training set. In contrast, the first two matching tasks from the semantic benchmark show a more pronounced performance drop, suggesting that exposure to a large number of samples is critical for learning high-quality semantic representations. Interestingly, performance on the clustering task remains strong even with limited data, consistently outperforming most other models. This indicates that our training scheme is particularly effective at structuring the embedding space semantically, even under data constraints.

Next, we evaluate the importance of also using paper titles and abstract sentences as positives instead of solely relying on the generated summaries. If just summaries are used, results display a slightly reduced performance on the SciRepEval benchmark and notable performance degradation on the semantic matching tasks, underscoring the importance of training on diverse input types that reflect the actual distribution of scientific text.

Third, we explore the effect of employing different similarity metrics as basis for the embedding space. While Euclidean distance is commonly used in embedding spaces for scientific text representation, many general-purpose embedding models instead rely on cosine similarity. To evaluate the effect of choosing one over the other, we also evaluate SemCSE using a cosine-based setting by applying a standard softmax-based contrastive loss (see Appendix \ref{app:abl_cosine}). The resulting model achieves nearly equivalent performance on the SciRepEval benchmark, although the distribution of results across task types shifts. Specifically, classification, proximity, and search tasks show modest improvements, while regression performance declines substantially. On our proposed semantic benchmark, on the other hand, the cosine-based model still displays state-of-the-art performance, but performs slightly worse compared to the Euclidean-based model, except for the query matching task where it performs better. Ultimately, these results validate the general effectiveness of leveraging matching LLM-generated summary pairs for injecting semantic understanding into the embedding model, regardless of the specific distance metric or loss function employed.

Finally, we investigate the role of using different but semantically related inputs by training a variant of our SemCSE model using the same sentence as both anchor and positive, thus mirroring the approach of unsupervised SimCSE. In this setup, positive pairs differ only due to dropout-induced variance. This strategy has shown strong performance on general semantic textual similarity benchmarks - primarily by improving embedding space uniformity (see Section \ref{sec:comp_analysis}) - so that our goal is to identify the contribution of distinct summaries in our method.

The results of this setup (denoted “Same Input”) confirm that even without distinct input pairs, the model achieves substantial improvements over the SciDeBERTa base model, validating the effectiveness of the overall learning objective when combined with triplet margin loss in Euclidean space. However, performance across both the semantic and SciRepEval benchmarks remains significantly below that of the full SemCSE model. This highlights the importance of training with semantically distinct yet related inputs: identical input pairs fail to provide the semantic variation necessary for robust representation learning. This distinction is further illustrated in Figure \ref{fig:embeddings}, which shows markedly stronger class separation in the embedding space when the full SemCSE objective is used.

These findings reinforce the analysis in Section \ref{sec:comp_analysis}, which emphasizes the advantages of using semantically diverse input pairs. By replacing SimCSE’s simple data augmentation strategy with a more meaningful signal, our approach yields a more challenging training task - ultimately leading to more effective and generalizable representations for the scientific domain.

\section{Conclusion}
\label{sec:conclusion}

In this work, we address the challenge of learning robust semantic embeddings for scientific texts. Recognizing the limitations of traditional citation-based supervision, we propose a paradigm shift towards a semantically-focused training and evaluation paradigm, resulting in the proposal of SemCSE and a novel benchmark for semantic evaluation.

While our proposed paradigm shift is grounded in existing literature that questions the semantic relatedness of papers linked by citations, we empirically demonstrate the improved semantic representation capabilities of SemCSE on several matching tasks and both quantify and visualize this enhanced semantic structuring of the underlying embedding space on a diverse clustering dataset.

Further, the analysis of the SciRepEval benchmark shows that our method especially excels at classification tasks, which benefit from a clear semantic separation in the embedding space and thus again demonstrates the benefits of the proposed training scheme.

Finally, our ablation studies further pinpoint the use of distinct yet semantically related summary pairs as a critical component of SemCSE's success, thus demonstrating the benefit of a semantically diversified training strategy in contrast to simple data augmentation.

Ultimately, the evidence presented strongly advocates for a paradigm shift towards semantically-oriented training and evaluation for scientific text embeddings, and we believe our novel evaluation scheme itself offers valuable insights into the capabilities of existing embedding models. Beyond the scientific domain, the core unsupervised training methodology of SemCSE holds promise as a broadly applicable strategy for learning high-quality embeddings across diverse fields.

\section{Limitations}

A core limitation of SemCSE is the reliance on LLM-generated summaries, which has the possibility of introducing systemic biases into the embedding model, and also poses a risk of learning incorrect representations in cases of hallucinations or factual errors.

Also, while SemCSE generates semantically meaningful embeddings, the interpretability of these embeddings remains a challenge. Understanding the exact influence that specific pieces of information within abstracts have on being nearby or far apart in the embedding space is not straightforward, which could limit the model's utility in applications where interpretability is crucial.

\section*{Acknowledgements}
This work was funded by Deutsche Forschungsgemeinschaft DFG (project
number 455913229; T.H., M.B., J.M.J., B.K-R, S.Z.).

\bibliography{bib}

\begin{thebibliography}{44}
\providecommand{\natexlab}[1]{#1}

\bibitem[{Achakulvisut et~al.(2016)Achakulvisut, Acuna, Ruangrong, and Kording}]{achakulvisut_science_2016}
Titipat Achakulvisut, Daniel~E. Acuna, Tulakan Ruangrong, and Konrad Kording. 2016.
\newblock \href {https://doi.org/10.1371/journal.pone.0158423} {Science concierge: A fast content-based recommendation system for scientific publications}.
\newblock 11(7):e0158423.

\bibitem[{Beltagy et~al.(2019)Beltagy, Lo, and Cohan}]{beltagy_scibert_2019}
Iz~Beltagy, Kyle Lo, and Arman Cohan. 2019.
\newblock \href {https://doi.org/10.18653/v1/D19-1371} {{SciBERT}: A pretrained language model for scientific text}.
\newblock In \emph{Proceedings of the 2019 Conference on Empirical Methods in Natural Language Processing and the 9th International Joint Conference on Natural Language Processing ({EMNLP}-{IJCNLP})}, pages 3615--3620. Association for Computational Linguistics.

\bibitem[{Bhagavatula et~al.(2018)Bhagavatula, Feldman, Power, and Ammar}]{bhagavatula_content-based_2018}
Chandra Bhagavatula, Sergey Feldman, Russell Power, and Waleed Ammar. 2018.
\newblock \href {https://doi.org/10.18653/v1/N18-1022} {Content-based citation recommendation}.
\newblock In \emph{Proceedings of the 2018 Conference of the North American Chapter of the Association for Computational Linguistics: Human Language Technologies, Volume 1 (Long Papers)}, pages 238--251. Association for Computational Linguistics.

\bibitem[{Bornmann et~al.(2021)Bornmann, Haunschild, and Mutz}]{bornmann_growth_2021}
Lutz Bornmann, Robin Haunschild, and Rüdiger Mutz. 2021.
\newblock \href {https://doi.org/10.1057/s41599-021-00903-w} {Growth rates of modern science: a latent piecewise growth curve approach to model publication numbers from established and new literature databases}.
\newblock 8(1):1--15.
\newblock Publisher: Palgrave.

\bibitem[{Brinner et~al.(2025)Brinner, Mustafa, and Zarrieß}]{brinner2025enhancingdomainspecificencodermodels}
Marc Brinner, Tarek~Al Mustafa, and Sina Zarrieß. 2025.
\newblock \href {https://arxiv.org/abs/2503.22006} {Enhancing domain-specific encoder models with llm-generated data: How to leverage ontologies, and how to do without them}.
\newblock \emph{Preprint}, arXiv:2503.22006.

\bibitem[{Chen et~al.(2025)Chen, Wang, Yang, Zhu, Zhao, Wei, and Dou}]{chen-etal-2025-little}
Haonan Chen, Liang Wang, Nan Yang, Yutao Zhu, Ziliang Zhao, Furu Wei, and Zhicheng Dou. 2025.
\newblock \href {https://doi.org/10.18653/v1/2025.naacl-long.64} {Little giants: Synthesizing high-quality embedding data at scale}.
\newblock In \emph{Proceedings of the 2025 Conference of the Nations of the Americas Chapter of the Association for Computational Linguistics: Human Language Technologies (Volume 1: Long Papers)}, pages 1392--1411, Albuquerque, New Mexico. Association for Computational Linguistics.

\bibitem[{Cohan et~al.(2020{\natexlab{a}})Cohan, Feldman, Beltagy, Downey, and Weld}]{cohan2020specter}
Arman Cohan, Sergey Feldman, Iz~Beltagy, Doug Downey, and Daniel Weld. 2020{\natexlab{a}}.
\newblock \href {https://doi.org/10.18653/v1/2020.acl-main.207} {{SPECTER}: Document-level representation learning using citation-informed transformers}.
\newblock In \emph{Proceedings of the 58th Annual Meeting of the Association for Computational Linguistics}, pages 2270--2282, Online. Association for Computational Linguistics.

\bibitem[{Cohan et~al.(2020{\natexlab{b}})Cohan, Feldman, Beltagy, Downey, and Weld}]{cohan_specter_2020}
Arman Cohan, Sergey Feldman, Iz~Beltagy, Doug Downey, and Daniel Weld. 2020{\natexlab{b}}.
\newblock \href {https://doi.org/10.18653/v1/2020.acl-main.207} {{SPECTER}: Document-level representation learning using citation-informed transformers}.
\newblock In \emph{Proceedings of the 58th Annual Meeting of the Association for Computational Linguistics}, pages 2270--2282. Association for Computational Linguistics.

\bibitem[{Frank and Afli(2024)}]{frank2024gasegenerativelyaugmentedsentence}
Manuel Frank and Haithem Afli. 2024.
\newblock \href {https://arxiv.org/abs/2411.04914} {Gase: Generatively augmented sentence encoding}.
\newblock \emph{Preprint}, arXiv:2411.04914.

\bibitem[{Gao et~al.(2019)Gao, He, Tan, Qin, Wang, and Liu}]{gao2019representation}
Jun Gao, Di~He, Xu~Tan, Tao Qin, Liwei Wang, and Tie-Yan Liu. 2019.
\newblock \href {https://arxiv.org/abs/1907.12009} {Representation degeneration problem in training natural language generation models}.
\newblock \emph{Preprint}, arXiv:1907.12009.

\bibitem[{Gao et~al.(2021)Gao, Yao, and Chen}]{gao2021simcse}
Tianyu Gao, Xingcheng Yao, and Danqi Chen. 2021.
\newblock \href {https://doi.org/10.18653/v1/2021.emnlp-main.552} {{S}im{CSE}: Simple contrastive learning of sentence embeddings}.
\newblock In \emph{Proceedings of the 2021 Conference on Empirical Methods in Natural Language Processing}, pages 6894--6910, Online and Punta Cana, Dominican Republic. Association for Computational Linguistics.

\bibitem[{Gao et~al.(2024)Gao, Xiong, Gao, Jia, Pan, Bi, Dai, Sun, Wang, and Wang}]{gao2024retrieval}
Yunfan Gao, Yun Xiong, Xinyu Gao, Kangxiang Jia, Jinliu Pan, Yuxi Bi, Yi~Dai, Jiawei Sun, Meng Wang, and Haofen Wang. 2024.
\newblock \href {https://arxiv.org/abs/2312.10997} {Retrieval-augmented generation for large language models: A survey}.
\newblock \emph{Preprint}, arXiv:2312.10997.

\bibitem[{Grattafiori et~al.(2024)}]{grattafiori2024llama3herdmodels}
Aaron Grattafiori et~al. 2024.
\newblock \href {https://arxiv.org/abs/2407.21783} {The llama 3 herd of models}.
\newblock \emph{Preprint}, arXiv:2407.21783.

\bibitem[{Günther et~al.(2023)Günther, Ong, Mohr, Abdessalem, Abel, Akram, Guzman, Mastrapas, Sturua, Wang, Werk, Wang, and Xiao}]{günther2023jina}
Michael Günther, Jackmin Ong, Isabelle Mohr, Alaeddine Abdessalem, Tanguy Abel, Mohammad~Kalim Akram, Susana Guzman, Georgios Mastrapas, Saba Sturua, Bo~Wang, Maximilian Werk, Nan Wang, and Han Xiao. 2023.
\newblock \href {https://arxiv.org/abs/2310.19923} {Jina embeddings 2: 8192-token general-purpose text embeddings for long documents}.
\newblock \emph{Preprint}, arXiv:2310.19923.

\bibitem[{Hadifar et~al.(2019)Hadifar, Sterckx, Demeester, and Develder}]{hadifar2019self}
Amir Hadifar, Lucas Sterckx, Thomas Demeester, and Chris Develder. 2019.
\newblock \href {https://doi.org/10.18653/v1/W19-4322} {A self-training approach for short text clustering}.
\newblock In \emph{Proceedings of the 4th Workshop on Representation Learning for NLP (RepL4NLP-2019)}, pages 194--199, Florence, Italy. Association for Computational Linguistics.

\bibitem[{Hjørland and Albrechtsen(1995)}]{hjorland_toward_1995}
Birger Hjørland and Hanne Albrechtsen. 1995.
\newblock \href {https://doi.org/10.1002/(SICI)1097-4571(199507)46:6<400::AID-ASI2>3.0.CO;2-Y} {Toward a new horizon in information science: Domain-analysis}.
\newblock 46(6):400--425.

\bibitem[{Hoffer and Ailon(2015)}]{triplet_loss}
Elad Hoffer and Nir Ailon. 2015.
\newblock Deep metric learning using triplet network.
\newblock In \emph{Similarity-Based Pattern Recognition}, pages 84--92, Cham. Springer International Publishing.

\bibitem[{Huang et~al.(2021)Huang, Tang, Zhong, Lu, Shou, Gong, Jiang, and Duan}]{huang2021whiteningbert}
Junjie Huang, Duyu Tang, Wanjun Zhong, Shuai Lu, Linjun Shou, Ming Gong, Daxin Jiang, and Nan Duan. 2021.
\newblock \href {https://doi.org/10.18653/v1/2021.findings-emnlp.23} {{W}hitening{BERT}: An easy unsupervised sentence embedding approach}.
\newblock In \emph{Findings of the Association for Computational Linguistics: EMNLP 2021}, pages 238--244, Punta Cana, Dominican Republic. Association for Computational Linguistics.

\bibitem[{Kanakia et~al.(2019)Kanakia, Shen, Eide, and Wang}]{kanakia_scalable_2019}
Anshul Kanakia, Zhihong Shen, Darrin Eide, and Kuansan Wang. 2019.
\newblock \href {https://doi.org/10.1145/3308558.3313700} {A scalable hybrid research paper recommender system for microsoft academic}.
\newblock In \emph{The World Wide Web Conference}, pages 2893--2899.

\bibitem[{Karpukhin et~al.(2020)Karpukhin, Oguz, Min, Lewis, Wu, Edunov, Chen, and Yih}]{karpukhin2020dense}
Vladimir Karpukhin, Barlas Oguz, Sewon Min, Patrick Lewis, Ledell Wu, Sergey Edunov, Danqi Chen, and Wen-tau Yih. 2020.
\newblock \href {https://doi.org/10.18653/v1/2020.emnlp-main.550} {Dense passage retrieval for open-domain question answering}.
\newblock In \emph{Proceedings of the 2020 Conference on Empirical Methods in Natural Language Processing (EMNLP)}, pages 6769--6781, Online. Association for Computational Linguistics.

\bibitem[{Kim et~al.(2023)Kim, Jeong, and Choi}]{10355927}
Eunhui Kim, Yuna Jeong, and Myung-Seok Choi. 2023.
\newblock \href {https://doi.org/10.1109/ACCESS.2023.3341612} {Medibiodeberta: Biomedical language model with continuous learning and intermediate fine-tuning}.
\newblock \emph{IEEE Access}, 11:141036--141044.

\bibitem[{Lee et~al.(2025)Lee, Roy, Xu, Raiman, Shoeybi, Catanzaro, and Ping}]{lee2025nvembed}
Chankyu Lee, Rajarshi Roy, Mengyao Xu, Jonathan Raiman, Mohammad Shoeybi, Bryan Catanzaro, and Wei Ping. 2025.
\newblock \href {https://arxiv.org/abs/2405.17428} {Nv-embed: Improved techniques for training llms as generalist embedding models}.
\newblock \emph{Preprint}, arXiv:2405.17428.

\bibitem[{Lee et~al.(2024)Lee, Dai, Ren, Chen, Cer, Cole, Hui, Boratko, Kapadia, Ding, Luan, Duddu, Abrego, Shi, Gupta, Kusupati, Jain, Jonnalagadda, Chang, and Naim}]{lee2024geckoversatiletextembeddings}
Jinhyuk Lee, Zhuyun Dai, Xiaoqi Ren, Blair Chen, Daniel Cer, Jeremy~R. Cole, Kai Hui, Michael Boratko, Rajvi Kapadia, Wen Ding, Yi~Luan, Sai Meher~Karthik Duddu, Gustavo~Hernandez Abrego, Weiqiang Shi, Nithi Gupta, Aditya Kusupati, Prateek Jain, Siddhartha~Reddy Jonnalagadda, Ming-Wei Chang, and Iftekhar Naim. 2024.
\newblock \href {https://arxiv.org/abs/2403.20327} {Gecko: Versatile text embeddings distilled from large language models}.
\newblock \emph{Preprint}, arXiv:2403.20327.

\bibitem[{Li et~al.(2020)Li, Zhou, He, Wang, Yang, and Li}]{li2020sentence}
Bohan Li, Hao Zhou, Junxian He, Mingxuan Wang, Yiming Yang, and Lei Li. 2020.
\newblock \href {https://doi.org/10.18653/v1/2020.emnlp-main.733} {On the sentence embeddings from pre-trained language models}.
\newblock In \emph{Proceedings of the 2020 Conference on Empirical Methods in Natural Language Processing (EMNLP)}, pages 9119--9130, Online. Association for Computational Linguistics.

\bibitem[{Li and Li(2024)}]{li2024aoe}
Xianming Li and Jing Li. 2024.
\newblock \href {https://doi.org/10.18653/v1/2024.acl-long.101} {{A}o{E}: Angle-optimized embeddings for semantic textual similarity}.
\newblock In \emph{Proceedings of the 62nd Annual Meeting of the Association for Computational Linguistics (Volume 1: Long Papers)}, pages 1825--1839, Bangkok, Thailand. Association for Computational Linguistics.

\bibitem[{Lv et~al.(2024)Lv, Niu, Han, and Li}]{lv_can_2024}
Hongjiang Lv, Zhibin Niu, Wei Han, and Xiang Li. 2024.
\newblock \href {https://doi.org/10.1007/s12650-024-01010-z} {Can {GPT} embeddings enhance visual exploration of literature datasets? {A} case study on isostatic pressing research}.
\newblock \emph{Journal of Visualization}.

\bibitem[{Meijer et~al.(2021)Meijer, Truong, and Karimi}]{meijer_document_2021}
H.~J. Meijer, J.~Truong, and R.~Karimi. 2021.
\newblock \href {https://doi.org/10.48550/arXiv.2107.05151} {Document embedding for scientific articles: Efficacy of word embeddings vs {TFIDF}}.
\newblock \emph{Preprint}, arxiv:2107.05151 [cs].

\bibitem[{{Mistral AI}(2025)}]{mistral}
{Mistral AI}. 2025.
\newblock \href {https://mistral.ai/news/mistral-small-3-1} {Mistral {Small} 3.1 {\textbar} {Mistral} {AI}}.

\bibitem[{Mysore et~al.(2022)Mysore, Cohan, and Hope}]{mysore2022multi}
Sheshera Mysore, Arman Cohan, and Tom Hope. 2022.
\newblock \href {https://doi.org/10.18653/v1/2022.naacl-main.331} {Multi-vector models with textual guidance for fine-grained scientific document similarity}.
\newblock In \emph{Proceedings of the 2022 Conference of the North American Chapter of the Association for Computational Linguistics: Human Language Technologies}, pages 4453--4470, Seattle, United States. Association for Computational Linguistics.

\bibitem[{Ostendorff et~al.(2022)Ostendorff, Rethmeier, Augenstein, Gipp, and Rehm}]{ostendorff_neighborhood_2022}
Malte Ostendorff, Nils Rethmeier, Isabelle Augenstein, Bela Gipp, and Georg Rehm. 2022.
\newblock \href {https://arxiv.org/abs/2202.06671 [cs]} {Neighborhood contrastive learning for scientific document representations with citation embeddings}.
\newblock \emph{Preprint}, arxiv:2202.06671 [cs].

\bibitem[{Pasternack(1969)}]{pasternack_scientific_1969}
Simon Pasternack. 1969.
\newblock \href {https://doi.org/10.1126/science.164.3880.669} {The scientific enterprise: Public knowledge. an essay concerning the social dimension of science. j. m. ziman. cambridge university press, new york, 1968. xii + 154 pp. cloth, \$3.95; paper, \$1.95.}
\newblock 164(3880):669--670.
\newblock Publisher: American Association for the Advancement of Science.

\bibitem[{Reimers and Gurevych(2019)}]{reimers2019sentence}
Nils Reimers and Iryna Gurevych. 2019.
\newblock \href {https://doi.org/10.18653/v1/D19-1410} {Sentence-{BERT}: Sentence embeddings using {S}iamese {BERT}-networks}.
\newblock In \emph{Proceedings of the 2019 Conference on Empirical Methods in Natural Language Processing and the 9th International Joint Conference on Natural Language Processing (EMNLP-IJCNLP)}, pages 3982--3992, Hong Kong, China. Association for Computational Linguistics.

\bibitem[{Singh et~al.(2023)Singh, D'Arcy, Cohan, Downey, and Feldman}]{singh_scirepeval_2023}
Amanpreet Singh, Mike D'Arcy, Arman Cohan, Doug Downey, and Sergey Feldman. 2023.
\newblock \href {https://arxiv.org/abs/2211.13308 [cs]} {{SciRepEval}: A multi-format benchmark for scientific document representations}.
\newblock \emph{Preprint}, arxiv:2211.13308 [cs].

\bibitem[{Sturua et~al.(2024)Sturua, Mohr, Akram, Günther, Wang, Krimmel, Wang, Mastrapas, Koukounas, Koukounas, Wang, and Xiao}]{sturua2024jina}
Saba Sturua, Isabelle Mohr, Mohammad~Kalim Akram, Michael Günther, Bo~Wang, Markus Krimmel, Feng Wang, Georgios Mastrapas, Andreas Koukounas, Andreas Koukounas, Nan Wang, and Han Xiao. 2024.
\newblock \href {https://arxiv.org/abs/2409.10173} {jina-embeddings-v3: Multilingual embeddings with task lora}.
\newblock \emph{Preprint}, arXiv:2409.10173.

\bibitem[{Subakti et~al.(2022)Subakti, Murfi, and Hariadi}]{subakti_performance_2022}
Alvin Subakti, Hendri Murfi, and Nora Hariadi. 2022.
\newblock \href {https://doi.org/10.1186/s40537-022-00564-9} {The performance of {BERT} as data representation of text clustering}.
\newblock \emph{Journal of Big Data}, 9(1):15.

\bibitem[{Tan et~al.(2023)Tan, Zhang, Zhao, and Zhang}]{tan_self-supervised_2023}
Shicheng Tan, Tao Zhang, Shu Zhao, and Yanping Zhang. 2023.
\newblock \href {https://doi.org/10.1007/s11192-023-04782-7} {Self-supervised scientific document recommendation based on contrastive learning}.
\newblock 128(9):5027--5049.

\bibitem[{Tang et~al.(2021)Tang, Sun, Jin, Wang, Zhang, and Wu}]{tang2021document}
Hongyin Tang, Xingwu Sun, Beihong Jin, Jingang Wang, Fuzheng Zhang, and Wei Wu. 2021.
\newblock \href {https://doi.org/10.18653/v1/2021.acl-long.392} {Improving document representations by generating pseudo query embeddings for dense retrieval}.
\newblock In \emph{Proceedings of the 59th Annual Meeting of the Association for Computational Linguistics and the 11th International Joint Conference on Natural Language Processing (Volume 1: Long Papers)}, pages 5054--5064, Online. Association for Computational Linguistics.

\bibitem[{Thirukovalluru and Dhingra(2025)}]{thirukovalluru-dhingra-2025-geneol}
Raghuveer Thirukovalluru and Bhuwan Dhingra. 2025.
\newblock \href {https://doi.org/10.18653/v1/2025.findings-naacl.122} {{G}en{EOL}: Harnessing the generative power of {LLM}s for training-free sentence embeddings}.
\newblock In \emph{Findings of the Association for Computational Linguistics: NAACL 2025}, pages 2295--2308, Albuquerque, New Mexico. Association for Computational Linguistics.

\bibitem[{Vasilyev et~al.(2025)Vasilyev, Sawaya, and Bohannon}]{vasilyev-etal-2025-preserving}
Oleg Vasilyev, Randy Sawaya, and John Bohannon. 2025.
\newblock \href {https://doi.org/10.18653/v1/2025.naacl-short.28} {Preserving multilingual quality while tuning query encoder on {E}nglish only}.
\newblock In \emph{Proceedings of the 2025 Conference of the Nations of the Americas Chapter of the Association for Computational Linguistics: Human Language Technologies (Volume 2: Short Papers)}, pages 321--341, Albuquerque, New Mexico. Association for Computational Linguistics.

\bibitem[{Wang et~al.(2023)Wang, Wang, Wang, Naidu, Bergen, and Paturi}]{NEURIPS2023_78f9c04b}
Jianyou~(Andre) Wang, Kaicheng Wang, Xiaoyue Wang, Prudhviraj Naidu, Leon Bergen, and Ramamohan Paturi. 2023.
\newblock \href {https://proceedings.neurips.cc/paper_files/paper/2023/file/78f9c04bdcb06f1ada3902912d8b64ba-Paper-Datasets_and_Benchmarks.pdf} {Scientific document retrieval using multi-level aspect-based queries}.
\newblock In \emph{Advances in Neural Information Processing Systems}, volume~36, pages 38404--38419. Curran Associates, Inc.

\bibitem[{Wang et~al.(2024)Wang, Yang, Huang, Yang, Majumder, and Wei}]{wang-etal-2024-improving-text}
Liang Wang, Nan Yang, Xiaolong Huang, Linjun Yang, Rangan Majumder, and Furu Wei. 2024.
\newblock \href {https://doi.org/10.18653/v1/2024.acl-long.642} {Improving text embeddings with large language models}.
\newblock In \emph{Proceedings of the 62nd Annual Meeting of the Association for Computational Linguistics (Volume 1: Long Papers)}, pages 11897--11916, Bangkok, Thailand. Association for Computational Linguistics.

\bibitem[{Wang and Isola(2020)}]{wang2022understand}
Tongzhou Wang and Phillip Isola. 2020.
\newblock Understanding contrastive representation learning through alignment and uniformity on the hypersphere.
\newblock In \emph{Proceedings of the 37th International Conference on Machine Learning}, ICML'20. JMLR.org.

\bibitem[{Wu et~al.(2020)Wu, Wang, Gu, Khabsa, Sun, and Ma}]{wu2020clearcontrastive}
Zhuofeng Wu, Sinong Wang, Jiatao Gu, Madian Khabsa, Fei Sun, and Hao Ma. 2020.
\newblock \href {https://arxiv.org/abs/2012.15466} {Clear: Contrastive learning for sentence representation}.
\newblock \emph{Preprint}, arXiv:2012.15466.

\bibitem[{Xu et~al.(2025)Xu, Chen, Wen, Liu, and He}]{xu-etal-2025-evaluating}
Borui Xu, Yao Chen, Zeyi Wen, Weiguo Liu, and Bingsheng He. 2025.
\newblock \href {https://aclanthology.org/2025.naacl-long.253/} {Evaluating small language models for news summarization: Implications and factors influencing performance}.
\newblock In \emph{Proceedings of the 2025 Conference of the Nations of the Americas Chapter of the Association for Computational Linguistics: Human Language Technologies (Volume 1: Long Papers)}, pages 4909--4922, Albuquerque, New Mexico. Association for Computational Linguistics.

\end{thebibliography}

\appendix

\section{Experimental Details}
\label{sec:appendix}

The code for training and evaluating our models, the best model checkpoint as well as the generated training data are available at \href{https://github.com/inas-argumentation/SemCSE}{github.com/inas-argumentation/SemCSE}.

\subsection{Dataset Creation}

The SciRepEval benchmark comprises six datasets that are used for training, available at \href{https://huggingface.co/datasets/allenai/scirepeval}{huggingface.co/datasets/allenai/scirepeval}. We randomly select subsets of each of these datasets as training data. The following datasets are used:
\begin{itemize}
    \item "mesh\_descriptors": 50,000 samples, medical domain
    \item "fos": 50,000 samples, various domains
    \item "search": 50,000 samples, various domains
    \item "same\_author": 50,000 samples, various domains
    \item "high\_influence\_cite": 50,000 samples, various domains
    \item "cite\_prediction\_new": 100,000 samples, various domains
\end{itemize}

The dataset used in the semantic evaluation benchmark uses 500 random samples from the train split of each of the datasets used in the SciRepEval benchmark: "relish", "high\_influence\_cite", "mesh\_descriptors", "biomimicry", "drsm", "cite\_prediction", "fos", "paper\_reviewer\_matching", "peer\_review\_score\_hIndex", "same\_author", "search", "tweet\_mentions".

More information on these datasets can be found in \cite{singh_scirepeval_2023}.

\subsection{Summary Generation}
\label{sec:data_generation}
We use Llama-3-8B \cite{grattafiori2024llama3herdmodels} to generate summarizing sentences for our dataset of scientific paper titles and abstracts. To this end, we append one of the following prompts to the abstract and let the LLM generate a continuation:
\begin{itemize}
    \item \textit{To summarize, the key findings of our research, stated in one sentence that includes all relevant information, are that}
    \item \textit{In summary, our research is concerned with}
    \item \textit{In summary, a comprehensive and detailed conclusive statement would be that}
    \item \textit{A comprehensive summary for our work would be that}
    \item \textit{The main takeaway from our work is that}
\end{itemize}

The development of this embedding approach started in a different domain, for which we experimented with using summaries generated by a chat LLM instead of using this text-continuation method. This did lead to worse results. We hypothesize that the sentences generated by the continuation-based method adhere more closely to the input data distribution, since the LLM effectively aims at continuing the abstract in the same style as before. Additionally, the task specification is less precise, since many continuations are possible. This could lead to a higher variance in the training set, which seems to be beneficial, as indicated by our ablations.

\definecolor{lightblue}{RGB}{50, 98, 168}

\begin{figure*}
    \centering
    \begin{tikzpicture}
        \begin{axis}[
            width=0.49\textwidth,
            height=0.35\textwidth,
            title={SciDeBERTa},
            xlabel={Principal Component},
            ylabel={Variance Explained (\%)},
            ylabel style={yshift=-0.3cm},
            xtick pos=left,
            ytick pos=left,
            ybar,
            bar width=2pt,
            ymin=0,
            ymax=32,
            xtick={1,10,20,30,40,50},
            ]
            \addplot[fill=lightblue, draw=none] table[x index=0, y index=1] {variance_data_SciDeBERTa.dat};
        \end{axis}
    \end{tikzpicture}
    \hfill
    \begin{tikzpicture}
        \begin{axis}[
            width=0.49\textwidth,
            height=0.35\textwidth,
            title={SemCSE},
            xlabel={Principal Component},
            ylabel={Variance Explained (\%)},
            ylabel style={yshift=-0.3cm},
            xtick pos=left,
            ytick pos=left,
            ybar,
            bar width=2pt,
            ymin=0,
            ymax=32,
            xtick={1,10,20,30,40,50},
            ]
            \addplot[fill=lightblue, draw=none] table[x index=0, y index=1] {variance_data_own.dat};
        \end{axis}
    \end{tikzpicture}
    \caption{Variance of the SciDocs MAG embedding space explained by the first 50 principal components for the base SciDeBERTa model and the trained SemCSE model.}
    \label{fig:variance_comparison}
\end{figure*}
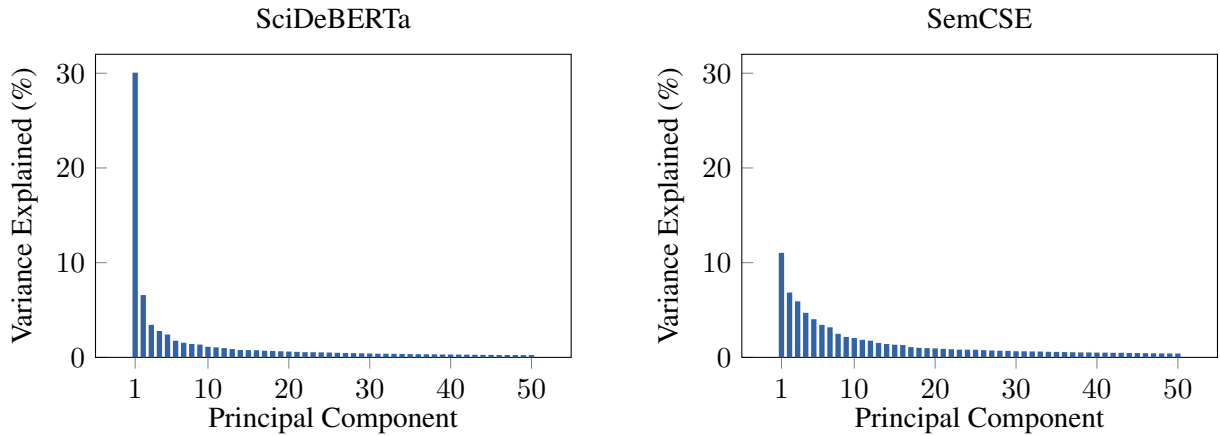

\subsection{Model Training}
We evaluate our SemCSE model after every 1000 batches, with each batch containing 32 pairs of summaries/abstract sentences.

We compute the evaluation scores on a set of 900 summary-abstract pairs. The evaluation uses the same ranking-based metric introduced in Section \ref{sec:generalization}, thus evaluating the average rank at which the correct match is retrieved.

We perform two different matching evaluations: One leverages two summaries per sample and aims at determining the matching summary score. The other takes the average embedding for both summaries and uses this to determine the matching title-abstract text.

If a new best score is achieved, the model is saved. Training is stopped as soon as the evaluation score did not decrease for 15 epochs.

We use an L2 regularizer applied to the embeddings of the anchors, which is averaged over all embeddings in the batch and weighted by a factor of $1/250$.

All ablations were trained with the same hyperparameter settings.

Hyperparameter search was performed by evaluating our model on training data for the training tasks described in \cite{singh_scirepeval_2023}.

The margin hyperparameter is set to 1, as is usual within most studies. This parameter does not have a notable effect, since the resulting embedding space can adhere to an arbitrarily large margin by simply scaling up the whole embedding space. Thus, precise values are not important, with only considerations being to select a value that does not cause numerical instabilities and reasonably fits to the magnitude of the embeddings of the pretrained model.

\subsection{SciRepEval Evaluation}

We evaluate SemCSE as well as the baselines on the SciRepEval benchmark \cite{singh_scirepeval_2023}. To calculate task-specific scores, we average over all metrics calculated for that task. In case a task includes full-dataset and few-shot results, all few-shot scores were averaged, and both few-shot and full-dataset results contributed evenly to the final scores.

Task-type averages were calculated by averaging over all task-specific scores for tasks of the respective type, while the complete-benchmark average was calculated by averaging over all scores for the individual tasks.

\subsection{Ablation: Cosine Similarity}
\label{app:abl_cosine}

We perform an ablation of SemCSE that uses cosine similarity as similarity measure instead of leveraging Euclidean distance. To this end, we use a loss formulation similar to SimCSE \cite{gao2021simcse}:

\begin{align*}
    \mathcal{L} = - \frac{1}{|\mathcal{B}|}\sum_{i \in \mathcal{B}} \log \frac{{\rm e}^{\text{sim}(e_{i, 1}, e_{i, 2}) / \tau}}{\sum_{j \in \mathcal{B}} {\rm e}^{\text{sim}(e_{i, 1}, e_{j, 2}) / \tau}}
\end{align*}

Here, $\text{sim}$ denotes the cosine similarity and $\tau$ is a temperature hyperparameter set to $0.07$, which is in line with similar work.

\section{Further Discussion}

\subsection{The Impact of Summary Quality}
\label{sec:summary_quality}
We use LLamA-3-8B to generate the summarizing sentences used for training. While this model performs reasonably well, it is surpassed by more recent and larger language models. This raises the question of whether the quality of generated summaries significantly affects the performance of our encoder model.

We argue that the quality of the summaries is not a substantial limiting factor, for several reasons:

\begin{enumerate}
\item Summarization is a relatively easy task for large language models, especially in the context of scientific abstracts, which are already highly condensed. Summarizing such short and structured texts requires minimal long-range reasoning or content synthesis, and even smaller models tend to perform well in this setting \cite{xu-etal-2025-evaluating}.

\item \textit{If} a model misinterprets a scientific abstract, the misunderstanding is likely to be systematic across all generated summaries. This consistency means the summaries still form valid training pairs, as the relationship between them remains meaningful even if they deviate slightly from the ground truth.

\item Minor factual inconsistencies between summaries do not invalidate their semantic relatedness. For instance, if one summary of a medical abstract states that a treatment was effective while another states it was not, both still pertain to the same core topic - the effectiveness of an intervention for this disease - and should be embedded similarly, since they would both be reasonable matches for a search query related to treatment outcomes regarding this disease.

\item Our preliminary experiments using chat-based LLMs (see Appendix \ref{sec:appendix}) actually resulted in lower performance. In these cases, continuations that were arguably lower in quality, but higher in variability, proved more effective for training. This suggests that overly polished summaries with reduced variance may not be optimal for learning robust semantic representations.

\item The query matching task is based on LLM-generated queries created by a significantly stronger LLM. Nevertheless, our results demonstrate that the SemCSE model can match these - likely more precise - queries to the corresponding abstracts, indicating that the model does not struggle with the different data distribution induced by the higher-quality model.

\end{enumerate}

While a more detailed evaluation of how different language models affect the performance of SemCSE would be valuable, generating 15 additional summaries for 350,000 input sentences would incur a substantial computational cost. Given the likely marginal insight this would provide, we consider such an investigation unjustifiable from a sustainability perspective.

\subsection{Analyzing Anisotropy}
\label{sec:anisotropy}

Contrastive embedding objectives have been shown to mitigate the issue of anisotropy \cite{gao2019representation, gao2021simcse} - the tendency for sentence embeddings to cluster within a narrow cone of the embedding space - an effect that has been shown to degrade the representational quality of embeddings \cite{li2020sentence}. Most prior analyses focus on hyperspherical embedding spaces induced by cosine similarity, where embeddings lie on the unit sphere and can only be distinguished by their angular separation. In this setting, a uniform distribution on the sphere is crucial for effective representation learning, as distances from the origin are no longer informative.

In Euclidean spaces, by contrast, it is theoretically possible for embeddings to occupy only a few dimensions while extending significantly along those axes to facilitate semantic disambiguation. However, in our experiments we apply L2 regularization, which encourages a compact embedding space and discourages extreme variance along any single dimension. As a result, we hypothesize that the contrastive objective - by pushing unrelated samples apart - promotes a more balanced use of the available dimensions, thereby reducing anisotropy even in Euclidean settings.

To empirically test this hypothesis, we embedded all samples from the SciDocs MAG dataset, applied mean-centering, and calculated the proportion of variance explained by each principal component. Unlike prior studies in hyperspherical spaces that examine the singular value distribution of the embedding matrix, this approach is more appropriate for Euclidean spaces, where the embedding matrix can be arbitrarily scaled, making raw singular values less interpretable.

The results, shown in Figure \ref{fig:variance_comparison}, compare the base SciDeBERTa model to our SemCSE-trained model. We observe a clear reduction in the dominance of the top principal components in the trained model, resulting in a smoother and more uniform variance distribution across components. This indicates a more isotropic embedding space and supports the conclusion that our training strategy improves the geometric quality of the learned representations.

\onecolumn
\section{SciRepEval Individual Results}
\label{app:scirep_individual}

\begin{table*}[h!]
  \centering
  \resizebox{\textwidth}{!}{%
  \begin{tabular}{l|ccc|ccc|ccc|c|c|c}
    \hline
    & \multicolumn{3}{c|}{\textbf{Biomimicry}} & \multicolumn{3}{c|}{\textbf{DRSM}} & \multicolumn{3}{c|}{\textbf{Fields of study}} & \textbf{MeSH} & \textbf{SD MAG} & \textbf{SD MeSH} \\
\textbf{Model} & F1 & F1 (fs-64) & F1 (fs-16) & F1 & F1 (fs-64) & F1 (fs-24) & F1 & F1 (fs-10) & F1 (fs-5) & F1 & F1 & F1 \\
    \hline \hline
    SciBERT & 73.37 & 37.27 & 16.00 & 76.84 & 56.31 & 46.05 & 40.02 & 30.07 & 21.61 & 76.71 & 79.50 & 79.99 \\
SciDeBERTa & 73.42 & 35.31 & 13.86 & 74.41 & 58.57 & 49.08 & 41.39 & 25.87 & 17.59 & 72.24 & 72.74 & 76.26 \\
SPECTER & 72.87 & 39.61 & 19.49 & \underline{77.34} & 61.07 & 48.88 & 42.43 & 32.98 & \underline{26.12} & 85.47 & 79.75 & 87.80 \\
SciNCL & 69.74 & 40.15 & \underline{21.26} & 74.73 & 61.24 & 49.68 & 44.14 & 32.76 & 25.00 & 86.17 & 81.11 & \underline{89.11} \\
SPECTER2 base & 74.21 & 39.20 & 14.49 & 76.42 & 55.68 & 43.31 & 42.21 & 25.56 & 15.68 & 86.76 & 81.03 & 89.00 \\
SPECTER2 proximity & 72.26 & 36.14 & 11.02 & 76.20 & 55.79 & 43.24 & 42.07 & 24.70 & 14.68 & 86.44 & 81.36 & 88.77 \\
\hline
jina-v2 & 69.98 & 0.63 & 0.00 & 75.46 & 47.05 & 36.52 & \textbf{47.14} & 24.68 & 11.08 & 86.18 & \underline{82.96} & 88.53 \\
jina-v3 & 71.96 & 17.07 & 0.64 & 77.15 & 55.49 & 41.62 & 45.19 & 24.74 & 10.86 & \underline{87.89} & 82.48 & 88.85 \\
RoBERTa SimCSE & 67.98 & \underline{40.88} & 16.85 & 76.29 & \textbf{66.67} & \textbf{57.60} & 46.01 & \textbf{34.48} & \textbf{26.30} & 82.60 & 80.46 & 84.04 \\
NvEmbed-V2 & \textbf{77.90} & 6.05 & 0.23 & 76.86 & 49.21 & 37.40 & \underline{46.04} & 17.37 & 4.12 & \textbf{89.47} & \textbf{84.68} & \textbf{90.60} \\
\hline
SemCSE (Ours) & \underline{77.21} & \textbf{42.58} & \textbf{21.47} & \textbf{78.00} & \underline{64.72} & \underline{54.55} & 43.31 & \underline{33.60} & 25.14 & 86.34 & 82.68 & 88.34 \\
\hline
  \end{tabular}
  }
  \caption{Results for Classification tasks on the SciRepEval benchmark. The best scores are bold, while second-best are underlined.}
  \label{tab:results_classification}
\end{table*}

\begin{table*}[h!]
  \centering
    \resizebox{\textwidth}{!}{%
  \begin{tabular}{l|ccccc}
    \hline
    & \textbf{Citation Count} & \textbf{Max hIndex} & \textbf{Peer Review} & \textbf{Publication Year} & \textbf{Tweet Mentions} \\
\textbf{Model} & Kendall's $\tau$ & Kendall's $\tau$ & Kendall's $\tau$ & Kendall's $\tau$ & Kendall's $\tau$ \\
    \hline
    \hline
SciBERT & \underline{39.59} & 17.19 & \textbf{23.37} & 30.87 & 25.67 \\
SciDeBERTa & 38.83 & \underline{17.40} & 21.29 & 32.80 & 24.69 \\
SPECTER & 35.38 & 15.51 & 18.12 & 30.12 & 27.73 \\
SciNCL & 34.71 & 15.00 & 20.03 & 30.02 & 26.34 \\
SPECTER2 base & 38.42 & 15.73 & 20.84 & \underline{35.57} & \underline{28.20} \\
SPECTER2 proximity & 38.58 & 14.56 & 20.22 & 33.65 & 27.22 \\
\hline
jina-v2 & 34.65 & 13.67 & 16.50 & 27.41 & 26.59 \\
jina-v3 & 34.46 & 15.28 & 17.43 & 30.39 & 26.65 \\
RoBERTa SimCSE & 36.37 & 11.59 & 14.89 & 27.30 & 24.62 \\
NvEmbed-V2 & \textbf{39.95} & \textbf{18.44} & 21.19 & \textbf{41.72} & \textbf{28.38} \\
\hline
SemCSE (Ours) & 38.90 & 17.14 & \underline{22.41} & 32.04 & 27.41 \\
\hline  \end{tabular}}
  \caption{Results for Regression tasks on the SciRepEval benchmark. The best scores are bold, while second-best are underlined.}
  \label{tab:results_regression}
\end{table*}

\begin{table*}[h!]
  \centering
  \resizebox{\textwidth}{!}{%
  \begin{tabular}{l|c|cccc|c|c|c|cc|cc|cc|cc}
    \hline
    & \textbf{H. Influence} & \multicolumn{4}{c|}{\textbf{Paper-Reviewer Matching}} & \textbf{RELISH} & \textbf{S2AND} & \textbf{Same Author} & \multicolumn{2}{c|}{\textbf{SD Cite}} & \multicolumn{2}{c|}{\textbf{SD CoCite}} & \multicolumn{2}{c|}{\textbf{SD CoRead}} & \multicolumn{2}{c}{\textbf{SD CoView}} \\
\textbf{Model} & MAP & P@10 h & P@5 h & P@10 s & P@5 s & nDCG & B3 F1 & MAP & MAP & nDCG & MAP & nDCG & MAP & nDCG & MAP & nDCG \\
    \hline
    \hline
SciBERT & 33.72 & 24.30 & 26.92 & 54.58 & 60.93 & 82.81 & 93.03 & 79.48 & 53.20 & 73.79 & 57.71 & 77.36 & 55.74 & 75.35 & 59.80 & 78.10 \\
SciDeBERTa & 31.85 & 24.21 & 26.73 & 53.46 & 60.00 & 81.90 & 92.13 & 75.28 & 45.94 & 69.13 & 50.01 & 72.24 & 49.27 & 71.11 & 53.18 & 73.96 \\
SPECTER & 42.89 & 25.51 & 33.27 & \textbf{56.17} & 65.79 & 90.07 & 93.12 & 86.53 & \underline{92.25} & 96.71 & 88.16 & 94.81 & 85.35 & 92.88 & 83.58 & 91.51 \\
SciNCL & 43.39 & 25.42 & 34.21 & 55.42 & 66.54 & 90.67 & \textbf{94.63} & 87.47 & \textbf{93.55} & \textbf{97.35} & \underline{91.66} & \underline{96.44} & \textbf{87.69} & \textbf{94.00} & 85.28 & 92.23 \\
SPECTER2 \small base & 44.96 & 25.42 & 34.02 & 55.51 & 66.73 & 91.63 & 93.00 & 87.00 & 91.97 & 96.69 & \textbf{91.70} & \textbf{96.56} & \underline{87.17} & \underline{93.71} & \textbf{85.52} & \textbf{92.50} \\
SPECTER2 \small prox. & \underline{46.07} & 25.61 & 34.21 & 55.61 & 66.17 & \underline{91.86} & 92.80 & \textbf{89.43} & 92.23 & \underline{96.84} & 91.13 & 96.28 & 86.85 & 93.53 & 85.18 & 92.26 \\
\hline
jina-v2 & 45.36 & \underline{25.70} & 34.39 & 55.51 & \textbf{67.66} & 90.76 & \underline{94.15} & 85.08 & 87.82 & 94.82 & 88.56 & 95.19 & 85.25 & 92.85 & 83.60 & 91.52 \\
jina-v3 & 45.40 & 25.33 & \textbf{34.77} & 55.23 & 66.54 & 91.60 & 94.03 & 85.24 & 87.38 & 94.61 & 87.32 & 94.59 & 84.73 & 92.57 & 83.48 & 91.40 \\
RoBERTa SimCSE & 41.37 & 25.23 & 29.91 & 54.77 & 63.74 & 87.61 & 93.23 & 80.49 & 76.48 & 88.88 & 79.08 & 90.28 & 76.39 & 88.14 & 78.70 & 89.05 \\
NvEmbed-V2 & \textbf{47.38} & \textbf{25.98} & \underline{34.58} & 55.42 & \underline{67.10} & \textbf{92.84} & 93.18 & 87.87 & 87.83 & 94.83 & 90.26 & 95.91 & 86.71 & 93.54 & \underline{85.38} & \underline{92.38} \\
\hline
SemCSE (Ours) & 44.35 & 25.42 & 32.34 & \underline{55.79} & 66.17 & 90.85 & 93.23 & \underline{88.66} & 87.24 & 94.39 & 88.84 & 95.20 & 85.39 & 92.85 & 84.09 & 91.74 \\
\hline  \end{tabular}
  }
  \caption{Results for Proximity tasks on the SciRepEval benchmark. The best scores are bold, while second-best are underlined.}
  \label{tab:results_proximity}
\end{table*}

\begin{table*}[h!]
  \centering
  \begin{tabular}{l|ccc}
    \hline
    & \textbf{NFCorpus} & \textbf{Search} & \textbf{TREC-CoVID} \\
    \textbf{Model} & nDCG & nDCG & nDCG \\
    \hline
    \hline
SciBERT & 53.34 & 71.49 & 79.73 \\
SciDeBERTa & 52.32 & 70.53 & 80.65 \\
SPECTER & 64.90 & 73.25 & 86.53 \\
SciNCL & 70.85 & 73.46 & 87.66 \\
SPECTER2 base & 72.03 & 73.76 & 89.46 \\
SPECTER2 proximity & 70.50 & 73.45 & 89.29 \\
\hline
jina-v2 & \underline{76.00} & 74.45 & 90.74 \\
jina-v3 & 75.12 & \underline{74.80} & \textbf{91.89} \\
RoBERTa SimCSE & 70.16 & 72.82 & 87.93 \\
NvEmbed-V2 & \textbf{81.47} & \textbf{75.18} & \underline{91.86} \\
\hline
SemCSE (Ours) & 72.53 & 73.32 & 89.82 \\
\hline  \end{tabular}
  \caption{Results for Ad-hoc Search tasks on the SciRepEval benchmark. The best scores are bold, while second-best are underlined.}
  \label{tab:results_search}
\end{table*}

\end{document}